\documentclass{article}
\usepackage{PRIMEarxiv}
\usepackage{times}
\usepackage{multirow}
\usepackage{amsmath}
\usepackage{amssymb}
\usepackage{nccmath}
\usepackage[numbers]{natbib}
\usepackage{hyperref}
\usepackage[T1]{fontenc}    
\usepackage{hyperref}       
\usepackage{url}            
\usepackage{booktabs}       
\usepackage{amsfonts}       
\usepackage{nicefrac}       
\usepackage{microtype}      
\usepackage{lipsum}
\usepackage{fancyhdr}       
\usepackage{graphicx}       
\graphicspath{{media/}}     
\newcommand{\abs}[1]{\left|#1\right|}

\title{Combining propensity score methods with variational autoencoders for generating synthetic data in presence of latent sub-groups
}

\author{Kiana Farhadyar\thanks{Corresponding author. Email: kiana.farhadyar@uniklinik-freiburg.de}\\
  Institute of Medical Biometry and Statistics,\\
  Faculty of Medicine and Medical Center, \\
  Freiburg Center for Data Analysis and Modeling\\
  University of Freiburg \\
  Freiburg, Germany\\
   \And
  Federico Bonofiglio  \\
  National Research Council of Italy, \\
  ISMAR \\ 
  La Spezia, Italy\\
   \And
  Maren Hackenberg \\
  Institute of Medical Biometry and Statistics,\\
  Faculty of Medicine and Medical Center, \\
  Freiburg Center for Data Analysis and Modeling,\\
  University of Freiburg \\
  Freiburg, Germany\\
   \And
  Daniela Zöller \\
  Institute of Medical Biometry and Statistics\\
  Faculty of Medicine and Medical Center, \\
    Freiburg Center for Data Analysis and Modeling,\\
  University of Freiburg \\
  Freiburg, Germany\\
   \And
  Harald Binder \\
  Institute of Medical Biometry and Statistics\\
  Faculty of Medicine and Medical Center, \\
  University of Freiburg \\
  Freiburg, Germany\\
}

\begin{document}
\maketitle

\begin{abstract}
In settings requiring synthetic data generation based on a clinical cohort, e.g., due to data protection regulations, heterogeneity across individuals might be a nuisance that we need to control or faithfully preserve. The sources of such heterogeneity might be known, e.g., as indicated by sub-groups labels, or might be unknown and thus reflected only in properties of distributions, such as bimodality or skewness. We investigate how such heterogeneity can be preserved and controlled when obtaining synthetic data from variational autoencoders (VAEs), i.e., a generative deep learning technique that utilizes a low-dimensional latent representation. To faithfully reproduce unknown heterogeneity reflected in marginal distributions, we propose to combine VAEs with pre-transformations. For dealing with known heterogeneity due to sub-groups, we complement VAEs with models for group membership, specifically from propensity score regression. The evaluation is performed with a realistic simulation design that features sub-groups and challenging marginal distributions. The proposed approach faithfully recovers the latter, compared to synthetic data approaches that focus purely on marginal distributions. Propensity scores add complementary information, e.g., when visualized in the latent space, and enable sampling of synthetic data with or without sub-group specific characteristics. We also illustrate the proposed approach with real data from an international stroke trial that exhibits considerable distribution differences between study sites, in addition to bimodality. These results indicate that describing heterogeneity by statistical approaches, such as propensity score regression, might be more generally useful for complementing generative deep learning for obtaining synthetic data that faithfully reflects structure from clinical cohorts.
\end{abstract}


\section{Introduction}
There has been a surge of interest in methods for generating synthetic datasets based on real clinical data \citep{goncalvesGenerationEvaluationSynthetic2020}. Such approaches may, e.g., be useful for providing data protection when even heavily sampled anonymized datasets do not meet privacy standards \citep{rocherEstimatingSuccessReidentifications2019}. In addition to the application for single datasets, another usage scenario is in federated computing platforms, such as DataSHIELD \citep{budin-ljosneDataSHIELDEthicallyRobust2015}, for simultaneously generating synthetic data at several sites and then pooling the synthetic data for test-driving analyses (e.g., \citealp{banerjeeDsSyntheticSyntheticData2022} or our own proposal in \citealp{lenzDeepGenerativeModels2021}). Beyond these data protection use cases, synthetic data can also be used for oversampling minority classes \citep{mullickGenerativeAdversarialMinority2019} or, more generally, augmenting the data (e.g., \citealp{antoniouDataAugmentationGenerative2018} and \citealp{saldanhaDataAugmentationUsing2022}). Furthermore, simulation studies and in silico clinical trials can benefit \citep{nowokSynthpopBespokeCreation2016,bollmannWhatCanReal2015,pappalardoSilicoClinicalTrials2019,zandDevelopmentSyntheticPatient2018}.

When using such techniques for clinical cohort data from observational studies, or also from randomized trials, faithful handling and potential preservation of heterogeneity across patients is important, in particular concerning sub-group structure. The importance of sub-groups in a clinical setting is reflected in a long history of research on biases that can arise when ignoring sub-group structure, e.g., as in Simpson's paradox \citep{simpsonInterpretationInteractionContingency1951}). Furthermore, there is a multitude of approaches for dealing with sub-group effects, such as propensity scores for properly assessing treatment effects \citep{rosenbaumCentralRolePropensity1983}, and also for more generally combining groups in clinical cohorts (e.g., our results in \citealp{zollerAutomaticVariableSelection2020}). Therefore, it might also be attractive to complement synthetic data techniques with approaches such as propensity score regression for handling heterogeneity due to known sub-groups. In addition to proposing a corresponding approach, we will also address heterogeneity due to unknown sub-groups. Contrary to the known sub-groups, which have explicit labels, the unknown ones are only reflected in marginal distributions. Therefore, we com,plement synthetic data techniques with pre-transformations to preserve the unknown structures and recover the bimodal or skewed distributions of continuous covariates.

The challenge of properly handling sub-groups already becomes apparent when considering one of the most prominent techniques for synthetic data generation, namely generative adversarial networks  \citep{goodfellowGenerativeAdversarialNets2014}), which had initially been developed for image data. There, the price for generating crisp synthetic images seems to be mode collapse, where certain sub-groups of the original dataset are no longer reflected \citep{goodfellowNIPS2016Tutorial2017}. Therefore, we consider an alternative popular technique as the basis for our proposed approach, specifically variational autoencoders (VAEs) \citep{kingmaAutoEncodingVariationalBayes2013}.  For modeling the relationships between multiple variables in a given dataset, VAEs build on an underlying low-dimensional latent representation, where artificial deep neural networks are used for estimating conditional distributions. The latter are amenable for combination with propensity scores obtained from regression models involving sub-group labels.
 
However, VAEs have also been developed with image data in mind, where some homogeneity in distributions can be assumed (\cite{nazabalHandlingIncompleteHeterogeneous2020}). This is reflected in an underlying assumption of a Gaussian prior on the latent representation, and thus VAEs have limitations with data deviating from unimodal symmetric distributions. 
While VAE-based approaches already exist for addressing data diverging from normal distributions based on modifying the prior on the latent representation (e.g., \citealp{guoVariationalAutoencoderOptimizing2020}), these are not flexible enough when different variables in the original data exhibit different kinds of peculiarities in their distribution. This motivates our pre-transformation component at the level of the original variables in our proposal.  

\begin{figure}[t]
\begin{center}
\includegraphics[width=1.0\textwidth]{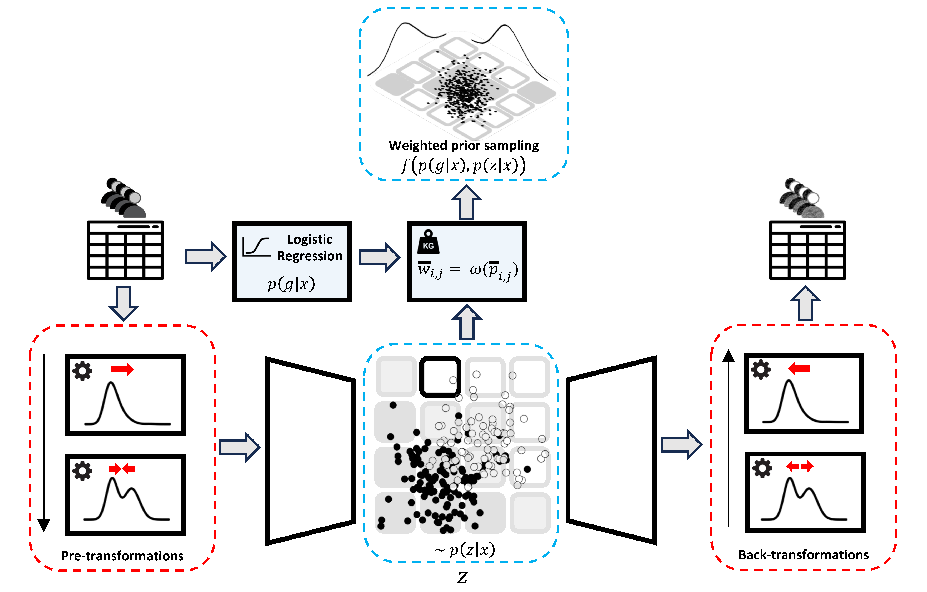}
\caption{Schematic overview of the proposed VAE-based approach, consisting of two primary components: (1) Unknown sub-groups within heterogeneous distributions are addressed through pre-transformations, as indicated by model components in red boxes. (2) Known sub-groups are handled using propensity score modeling \( p(g|x) \) and weighted prior sampling, indicated by blue boxes. The function $\omega$ incorporates the weights based on estimated propensity scores shown in Equations ~\ref{eq:commonclassindweighting} or ~\ref{eq:specificclassindweighting}.}

\label{fig:main_figure}
\end{center}
\end{figure}

There are also proposals for synthetic data outside the deep neural network community, e.g., using sampling based on the correlation matrix  \citep{koliopanosSimpletoUsePackageMimicking2023}. Similarly, we have introduced an approach based on Gaussian copula together with simple non-disclosive summaries \citep{bonofiglioRecoveryOriginalIndividual2020}. While we will consider the latter for performance comparison, our focus is on VAEs as their latent representations provide a starting point for complementing information from propensity score approaches. Figure ~\ref{fig:main_figure} shows the schematic overview of our approach.
Section ~\ref{sec:methods} introduces the proposed approach, specifically highlighting how heterogeneity due to known and unknown sub-group structures is handled. Section ~\ref{sec:marginal-results} contrasts our pre-transformation-enhanced VAE with other techniques in a simulation study and real data from a stroke trial. Section ~\ref{sec:propensity-score-reults} presents the combination of propensity scores with the latent representation of VAEs for simulation data, and weighted sampling is illustrated for the stroke trial example. We conclude with a discussion in Section ~\ref{discussion}. Source code for our approach is available on \href{https://github.com/kianaf/LatentSubgroups}{GitHub}.

\section{Methods} \label{sec:methods}

\subsection{General Framework}\label{subsec:general-framework}

Encoding data into a latent space allows for a better understanding of the data structure by revealing patterns not apparent in the original high-dimensional space. Specifically, dimension reduction techniques using two or three dimensions provide a visual insight into the data. If we consider the latent space to be given by a random variable \( z \), we can define a model \( p(z|x) \) as the probability distribution of the latent space given \( x \), which denotes the whole set of observations. As mentioned before, we should consider the heterogeneity between sub-groups in the dataset. In instances where we have known sub-groups with labels, a distinct model can approximate \( p(g|x) \), where \( g \) is a random variable producing the sub-group label. In this setting, our objective is to formulate a function \( f(p(z|x), p(g|x)) \) so that we can produce the structure of interest in our generated data (e.g., removing systematic differences or pronouncing one sub-group structure). For the other sub-groups with no explicit label, i.e., where we do not have access to \( p(g|x) \), and \(f\) would be implemented only based on \(p(z|x)\), our goal is to shape a latent structure reflecting the unknown sub-groups and consequently reconstruct the marginal distributions which are indications of existing underlying, yet unrecognized, sub-group structures. In the following sections, we explain different parts of our general framework, including the approximation for \(p(z|x))\) using a variational autoencoder, dealing with unknown and known sub-groups and how we implement \(f\). 

\subsection{Variational autoencoders (VAEs)}

One of the standard methods to approximate \( p(z|x) \) is the use of a specific type of autoencoders called variational autoencoder (VAE). Autoencoders consist of an encoder and a decoder, which are both multi-layer perceptrons, i.e., a neural network with one input layer, one output layer, and one or multiple hidden layers. As shown in Equation ~\ref{eq:nnlayer}, each layer, denoted by $l$, corresponds to a linear combination of its inputs $ h^{l-1}$ (which is the output vector of the previous layer or input data if $l = 1$) and weights $w^{(l)}$, and biases $b^{(l-1)}$, followed by a non-linear transformation $g^{(l)}$ called activation function. The layer output is a vector, which is used as input for the next layer.

\begin{align}
    \mathbf{h}^{(l)} = g^{(l)}\left(\mathbf{W}^{(l)}\mathbf{h}^{(l-1)} + \mathbf{b}^{(l)}\right)
    \label{eq:nnlayer}
\end{align}

The encoder part reduces the dimensions of the input layer to a latent embedding, and the decoder part tries to reconstruct the data from that \citep{rumelhartLearningRepresentationsBackpropagating1986}. A VAE \citep{kingmaAutoEncodingVariationalBayes2013} is a probabilistic version of an autoencoder, where the latent representation is considered to be given by a random variable with a prior distribution assumed to be a standard normal distribution. Based on Bayes' rule, the posterior distribution of the latent variable  \( z \) given the observed variable  \( x \) can be obtained via the Bayes' rule in Equation ~\ref{eq:bayesrule}. The integral in the denominator of the formula is computationally intractable, even for a relatively low-dimensional $z$. One solution is to use variational inference to approximate $p(z|x)$ by a distribution $q(z|x)$, which is a member of a parametric family of distributions, e.g., a Gaussian distribution with diagonal covariance, which is typically used in VAEs. Then, finding the posterior becomes an optimization problem, i.e., minimizing the Kullback-Leibler (KL) divergence between these two distributions, which can be calculated as shown in Equation ~\ref{eq:kld}.

\begin{equation}
p(z|x)=\frac{p(z,x)}{p(x)}=\frac{p(z,x)}{\int p(z,x) dz}
\label{eq:bayesrule}
 \end{equation}

\begin{equation}
\begin{aligned}
D_{K L}(q \| p)=E_{q(z|x)}[\log (q(z|x))-\log (p(x, z))]+\log \left(p(x)\right)
\end{aligned}
\label{eq:kld}
 \end{equation}

Since $\log \left(p_{x}(x)\right)$ in Equation ~\ref{eq:kld} is constant, and the KL divergence is a positive value, minimizing it is equivalent to maximizing the so-called evidence lower bound (ELBO) $E_{q(z|x)}[\log (q(z|x))-\log (p(x, z))]$. In the VAE, the encoder models $q_{\varphi}(z |x)$, where the parameters $\varphi$ are the encoder weights and biases, and the decoder models $p_{\theta}(x |z)$, with parameters $\theta$. The ELBO can be rewritten to obtain the loss function, shown in Equation ~\ref{eq:loss}, optimizing $\varphi$ and $\theta$. The first term on the right-hand side corresponds to reconstruction loss. The second term is the Kullback-Leibler divergence between the approximated posterior and the prior distribution. 

 \begin{equation}
\operatorname{loss}\left(x_{i}\right)=-E_{q_{\varphi}\left(z|x_{i}\right)}\left[\log p_{\theta}\left(x_{i}|z\right)\right]+ D_{K L}\left(q_{\varphi}\left(z|x_{i}\right)|| p_{\theta}(z)\right)
\label{eq:loss}
 \end{equation}

There are two methods for obtaining synthetic data from a trained VAE: 1) sampling $z$ from the approximated posterior given the original data or 2) sampling $z$ from the standard normal distribution (prior), followed in both cases by using the obtained values of $z$ as input for the decoder. The latter can better preserve the privacy because the original data can influence the synthetic data only via the trained parameters of the decoder. Therefore, if the VAE is not overfitted, having a low-dimensional latent space and sampling from prior can decrease the risk of data leakage.

\subsection{Dealing with unknown sub-groups}
To preserve the unknown sub-group structure, we aim to faithfully recover the marginal distributions. In this work, we concentrate on reconstructing Bernoulli (for binary variables), skewed and bimodal distributions. First, we need a VAE architecture to generate both continuous and binary variables (Section ~\ref{subsubsec:minor-modification}). Then, we use pre-transformations to transform the original data to remove skewness and bimodality so that a VAE can better model it. As is common in machine learning, to speed up the VAE training, we scale the data between zero and one. This needs to be considered in the backward process as well, i.e., after getting the output from the VAE, we have to transform the output back. Figure ~\ref{fig:main_figure} shows how the pre-transformation steps are incorporated into the general framework. The pre-transformation for skewed distributions is explained in Section ~\ref{subsubsec:boxcox}, and the pre-transformation for bimodal distributions is described in Section ~\ref{subsubsec:power}.

\subsubsection{VAE for combining continuous and binary variables} \label{subsubsec:minor-modification}

To generate both continuous and binary variables, we use an architecture with separate parts corresponding to the two variable types. For a decoder with $l+1$ layers, hidden layer $h_{D}^{(l)}$ serves as the joint basis for continuous and binary covariates, e.g., for representing correlation patterns between the two types of variables. Then, in the next layer, we have a group of neurons denoted by $\mu_{D}$ and $\sigma_{D}$ for the continuous variables and a group of neurons represented by $\pi_{D}$ for the binary variables. This means that the reconstructed values for the continuous variables are subsequently sampled from $N\left(\mu_{D}(h_{D}^{(l)}(z)), \sigma_{{D}}(h_{D}^{(l)}(z))\right)$ and for the binary variables by sampling from Bernoulli $(\pi_{D}(h_{D}^{(l)}(z)))$. Assuming $x_{i, j}$ as the \( j \)-th continuous variable of $x_{i}$ and $x_{i, k}$ as the \( k \)-th binary variable of $x_{i}$ and for \( x \) with $p_{c}$ continuous variables and $p_{b}$ binary variables, the loss function can be computed by Equation ~\ref{eq:totalloss}. The parameters of VAE are the weights and biases of the encoder and decoder $(\varphi, \theta)$.

 \begin{equation}
 \begin{aligned}
    \operatorname{loss}\left(x_{i}\right)= & -\sum_{k=1}^{p_{b}} \operatorname{logpdf} \left(\operatorname{Bernouli}\left(\pi_{D_{k}}(h_{D}^{(l)}(z_{i}))\right), x_{i, k}\right)\\
    & -\sum_{j=1}^{p_{c}} \operatorname{logpdf}\left(\operatorname{Normal}\left(\mu_{D_{j}}(h_{D}^{(l)}(z_{i})),\sigma_{D_{j}}(h_{D}^{(l)}(z_{i}))\right), x_{i, j}\right) \\
    & +D_{K L}\left(q_{\varphi}\left(z|x_{i}\right) \| p_{\theta}(z)\right).
\end{aligned}
\label{eq:totalloss}
 \end{equation}

With this modification, we can generate both binary and continuous variables, and thus cover the heterogeneity in the data types. However, another well-known problem when having different data types, e.g., binary and continuous or different sources, e.g., image and tabular data, is data fusion. There are different ways to tackle this (introduced in \cite{stahlschmidtMultimodalDeepLearning2022}). Given that there is no universally superior method for data fusion, we try two strategies as a hyper-parameter in each of our experiments. In this work, we investigate the early fusion, i.e., concatenation of the data type from the beginning, or late fusion, i.e., having two different encoders for binary and continuous variables.

\subsubsection{Box-Cox transformation}\label{subsubsec:boxcox}

To remove the skewness, we use a family of power transformations called the Box-Cox transformation, shown in Equation ~\ref{eq:boxcox}, suggested by \citealp{boxAnalysisTransformations1964}. In this formula, $\lambda_{2}$ is a shifting value to make the data positive, and $\lambda_{1}$ is the main parameter of the transformation. To estimate $\lambda_{1}$, we minimize the negative log-likelihood (Equation ~\ref{eq:negativeLL}) of the transformed values using gradient descent. In this optimization problem, we try to find the local minimum of this criterion using the gradient concept. After getting the output from the VAE, we have to transform the data back. The back-transformation for Box-Cox transformation is shown in Equations ~\ref{eq:boxcox-reverse}. 

 \begin{equation}
 \begin{aligned}
f_{\text {BoxCox}}\left(x ; \lambda_{1}, \lambda_{2}\right)= \begin{cases}\frac{\left(x+\lambda_{2}\right)^{\lambda_{1}}-1}{\lambda_{1}} & \lambda_{1} \neq 0 \\ \ln \left(x+\lambda_{2}\right) & \lambda_{1}=0\end{cases}
\label{eq:boxcox}
\end{aligned}
 \end{equation}

\begin{equation}
\begin{aligned}
 L(\lambda_1, \lambda_2 | x) = -\frac{N}{2} \log(\sigma^2 + \epsilon) & + (\lambda_1 - 1) \sum_{i=1}^{N} \log(x_i + \lambda_2 + \epsilon) \\
& \text{where}\\
\quad \sigma^2 =   \text{Var} (& f_{\text {BoxCox}}\left(x ; \lambda_{1}, \lambda_{2}\right))
\label{eq:negativeLL}
\end{aligned}
\end{equation}

 \begin{equation}
 \begin{aligned}
f^{-1}_{\text{BoxCox}}(y) = 
\begin{cases} 
\sqrt[\lambda_1]{\lambda_1 y^{(\lambda)} + 1} - \lambda_2, & \lambda_1 \neq 0 \\
e^{y(\lambda)} - \lambda_2, & \lambda_1 = 0 
\end{cases}
\label{eq:boxcox-reverse}
\end{aligned}
 \end{equation}
 
\subsubsection{Transformation for bimodality}\label{subsubsec:power}

The second transformation aims to make a bimodal distribution closer to an unimodal one by bringing the peaks closer and keeping the shape of the tails close to a normal distribution. We use a power function $x^{\rho}$ with the odd integer $(\rho=2 k+1$ with $k=1,\ldots, N)$, and this will work if we can shift and scale values such that the two peaks of the bimodal distribution fall within $(-1,1)$. Therefore, we must find the best values for the shifting parameter $\alpha$, positive scaling parameter $\beta^2$, and power $\rho$. to be able to continuously differentiate w.r.t these parameters for gradient-based optimization, we use $\operatorname{sgn}(x)|x|^{\rho}$, so that we have the same behavior for all different values of $\rho$. Hence, our transformation will be as Equation ~\ref{eq:power-tr}.

\begin{equation}
f(x) =  \operatorname{sgn}(\frac{(x + \alpha)}{\beta^2} ) |\frac{(x + \alpha)}{\beta^2} |^{\rho}
\label{eq:power-tr}
\end{equation}

For parameter optimization, we need a criterion that reflects closeness to an unimodal distribution. We considered maximum likelihood and the bimodality coefficient ($\left(b=\frac{\gamma^{2}+1}{\kappa}\right.$  where $\gamma$ is the skewness and $\kappa$ is the kurtosis), which both did not give adequate results as they decreased the variance too strongly. Therefore, we minimize a 1-sigma criterion (shown in Equation ~\ref{eq:1sigmarule}) to optimize the parameters. In this Equation, $Q_\tau(x)$ represents the $\tau$-th percentile of $x$. This optimization problem requires careful initialization of the parameters since the 1-sigma criterion is only a proxy for the deviation from an unimodal distribution. First, to have $\rho > 1$, we define the power parameter as $\rho = 1 + pow^2$. We start with $pow = 0$ and $\beta ^ 2 = 1$ for the scaling parameter to keep the data unchanged if it is normal/unimodal. Furthermore, finding the valley between two peaks in a heuristic way is a good starting point for the shifting variable ($\alpha$). We use an iterative heuristic algorithm based on kernel density estimation to initialize this value. In this method, we start estimating the density function with a very small bandwidth and find the local maxima of the function. Consequently, we gradually increase the bandwidth and continue until we only have a limited number of peaks (e.g., five). Then, we pick the two highest peaks and the deepest valley between these two. The value of the valley can be set as the initial value of $\alpha$.

\begin{equation}
    \begin{aligned}
    1-\operatorname{sigma}_{\text{criterion}}(x) & = \left| Q_{0.84}(x) - Q_{0.5}(x) - \sigma_{x} \right| + \left| Q_{0.5}(x) - Q_{0.16}(x) - \sigma_{x} \right|
    \end{aligned}
    \label{eq:1sigmarule}
\end{equation}

Like the first transformation, we also need the back transformation function for the second one. The reverse function is shown in Equation ~\ref{eq:power-reverse}. Applying the pre-transformations addresses the challenge of heterogeneity in the distributions of continuous variables.
\begin{equation}
f^{-1}(y) = \beta^2 y^{\frac{1}{\rho}} \operatorname{sgn}(y) - \alpha
\label{eq:power-reverse}
\end{equation}

\subsection{Dealing with known sub-groups}

\subsubsection{Propensity score estimation}\label{subsubsec:ps}
Dealing with known sub-groups requires an approach that generates the structure of interest, i.e., removing the systematic differences between sub-groups or pronouncing the characteristics specific to one sub-group. To sample from areas of the latent space which have our structure of interest, we need a quantitative guide, such as \( p(g|x) \), to be used as a weighting system when sampling from the prior distribution of \( z \). Therefore, we build a model for estimation of \(p(g|x)\) to predict the sub-group membership for each observation \(x_i\). To achieve this, we use propensity scores, i.e., the probability of an observation belonging to a group given a set of covariates \cite{liBalancingCovariatesPropensity2018}. We use a logistic regression for the binary classification of sub-groups in our datasets outlined in Section ~\ref{subsec:datasets}. The model prediction for a data point $x_i$ will then be the probability of \( x_i \) belonging to the sub-group number one, i.e., \(p(g=1|x = x_i)\), where \( g \in \{1, 2\}\). We use the original data space for propensity score estimation for two main reasons. First, this approach allows us to leverage the rich information inherent in the original data, which can be crucial for accurate propensity score estimation. Second, logistic regression is effective in the original data space, even when faced with complex data distributions. This model robustness might not hold in a reduced-dimensional latent space. In such lower-dimensional spaces, the simplification of data can lead to a loss of important information, especially when dealing with heterogeneous and complex distributions. 

\subsubsection{Propensity score-based sampling method} \label{subsec:psweightedsampling}
We can use the propensity score concept for assigning weights to different areas of the latent space learned by the VAE. To see whether we can use propensity scores concept as a guide for sampling from the latent space, we visualize the latent space with the propensity score values. For this, we divide the latent space into a grid of cells with a tenable size $d$. For a two-dimensional latent space $z=((z)_1,(z)_2)$, we then define a matrix $A$, where $A_{i,j}$ is a cell in the grid on the latent space. In this grid, $i$ denotes the index of the cell along the $(z)_1$-axis, while $j$ denotes the index of the cell along the $(z)_2$-axis. 
Therefore, for each cell $A_{i,j}$ we have that for all $z\in A_{i,j}$: 

\begin{equation}
    \begin{aligned}
    &\min_z(z)_1 < (z)_1 < (\min_z(z)_1 + (i-1) \cdot d), 
    \quad \text{for } i = 1, \ldots, N_1 = \left\lceil \frac{\max_z(z)_1 - \min_z(z)_1}{d} \right\rceil, \\
    &\min_z(z)_2 < (z)_2 < (\min_z(z)_2 + (j-1) \cdot d), 
    \quad \text{for } j = 1, \ldots, N_2 = \left\lceil \frac{\max_z(z)_2 - \min_z(z)_2}{d} \right\rceil. \\
    \end{aligned}
\end{equation}

After making a grid on the latent space, we need to calculate the propensity score for each cell. For this, we fit a logistic regression on the observations $x_{1 \ldots n}$, and then we calculate the propensity score using the predictions of a logistic regression model. After this, each point in the latent space of VAE $z_k$, which is the mapping of an observation $x_k$, has a propensity score $p_{x_k}$. Then, we calculate the propensity score for each cell, averaging the propensity score of the points that are in that specific cell. This is shown in Equation ~\ref{eq:meanps}.

\begin{equation}
    \begin{aligned}
    \bar{p}_{i,j} = \frac{1}{n} \sum_{k=1}^{n} p_{x_k} \cdot \mathbb{I}(z_k \in A_{i,j})
    \end{aligned}
    \label{eq:meanps}
 \end{equation}

After the propensity score calculation, as shown in Figure ~\ref{fig:main_figure} we can overlay the grid with the scatter plot of the latent space, color-coded by the existing sub-groups in our dataset. If the cells in the grid, colored by propensity score, correspond to the color of the majority group of the points in each cell, we can use this as a guide for sampling from the prior distribution, i.e., we can define weights based on the scenario in which we want to generate synthetic data. For this, we use Inverse Probability of Treatment weighting (IPTW) \citep{austinMovingBestPractice2015} to define a new weighting approach for our scenario. Suppose we only want to generate individuals that are common for both sub-groups. In that case, we can use Equation ~\ref{eq:commonclassindweighting}. If we want only to have individuals with the characteristics of one group, say, where $g=0$, and individuals should have a small value of $\bar{p}_{i,j}$, we can use the weighting system shown in Equation~\ref{eq:specificclassindweighting}. In both equations, $\delta$ denotes the acceptable deviation from $\bar{p}_{i,j} = 0.5 $, representing the areas common for both populations.  

\begin{equation}
    \begin{aligned}
    w_{i,j} = 
    \begin{cases} 0   & \abs {\bar{p}_{i,j} - 0.5} > \delta \\ 
    \frac{1}{\bar{p}_{i,j}} & \bar{p}_{i,j}>0.5 \\
    \frac{1}{1 - \bar{p}_{i,j}} & \bar{p}_{i,j}<0.5 
    \end{cases}
    \end{aligned}
    \label{eq:commonclassindweighting}
 \end{equation}

\begin{equation}
    \begin{aligned}
    w_{i,j} = 
    \begin{cases} 0   & \bar{p}_{i,j} > 0.5 + \delta \\ 
    \frac{1}{\bar{p}_{i,j}} & \bar{p}_{i,j} \leq 0.5 + \delta 
    \end{cases}
    \end{aligned}
    \label{eq:specificclassindweighting}
 \end{equation}

We obtain weights for each cell using Equation ~\ref{eq:commonclassindweighting} or Equation ~\ref{eq:specificclassindweighting} and subsequently normalize them, as shown in Equation ~\ref{eq:normalizedweights}.

\begin{equation}
    \begin{aligned}
    \bar{w}_{i,j} = \frac{w_{i,j}}{\sum_{i=1}^{N_1} \sum_{j=1}^{N_2} w_{i,j}} 
    \end{aligned}
    \label{eq:normalizedweights}
 \end{equation}

We can now use propensity scores as a guide for prior sampling to generate a synthetic dataset. Specifically, we sample from the prior distribution, $N(0, 1)$, then find the corresponding cell and the weight assigned to that cell. Next, we sample from a Bernoulli distribution with $P(X = 1) = \bar{w}_{i,j}$. This step is the decision flag to include or reject the sampled value. Repeating the process until we reach the intended sample size gives us a set of samples to feed to the decoder and get the output as our synthetic dataset.

\section{Evaluation of the method for unknown sub-group structures} \label{sec:marginal-results}

 \subsection{Baseline approaches}
 To evaluate our method in dealing with unknown sub-groups, we compare the utility of synthetic data generated with the standard VAE \cite{kingmaAutoEncodingVariationalBayes2013} with the minor modifications in encoder and decoder (without pre-transformations), the VAE with an autoregressive implicit quantile network (AIQN) \citep{ostrovskiAutoregressiveQuantileNetworks2018} (called QVAE here), generative adversarial networks (GAN) \citep{goodfellowGenerativeAdversarialNets2014} and NORTA-J, our Gaussian copula-based approach with first four moments \citep{bonofiglioRecoveryOriginalIndividual2020}. For simplicity, we use the same architecture for all the VAE-based approaches. The evaluation metrics are explained in Section ~\ref{subsec:comparisonapproaches}.

In the QVAE approach, quantile regression allows for more flexibility in the VAE latent space. Specifically, a neural network embedded in the latent space implements the quantile regression for each dimension. For each data point $x_i$, we get a $z_i$ in the latent space. We use a random number $0.05<$ $\tau<0.95$ as an input of each quantile network, and then for each dimension $k$, we use the $\tau$ and $z_{i_1}, \ldots, z_{i_{k-1}}$ as the input and $z_{i_k}$ as the output. This means that for the first dimension, the network has one input, i.e., $\tau$, and one output, i.e., $z_1$. Because we have a conditional network based on the previous dimensions and $\tau$, we need to use the best order of $z_{i_{1, \ldots, l}}$ for the quantile network architecture. We use the Kolmogrov-Smirnov test to determine which order makes the conditional distribution closer to a normal distribution. We train the network with the quantile regression loss function for each dimension. For more information on the details of the QVAE approach, see \cite{ostrovskiAutoregressiveQuantileNetworks2018}.

GANs comprise two multiple-layer perceptrons, called the discriminator and the generator. The generator part is responsible for generating synthetic data, and the discriminator aims to distinguish between real data and generated data. The better the generator, the harder it is for the discriminator to distinguish real and generated data. After training the generator to fool the discriminator, which is trained simultaneously, the generator should be able to generate realistic synthetic data. For more information see \cite{goodfellowGenerativeAdversarialNets2014}.

The method from our previous work, which we call NORTA-J here, infers the original individual person data (IPD) characteristics from summary statistics. This method generates synthetic data through a Gaussian copula inversion technique known as NORTA, which models the dependency structure of the data variables. The marginal distributions of IPD are constructed using the Johnson system of distributions, parameterized by empirical marginal moments (e.g., mean, variance) and the correlation matrix \citep{bonofiglioRecoveryOriginalIndividual2020}.

 \subsection{Approaches for comparison and evaluation criteria} \label{subsec:comparisonapproaches}

As the first quantitative measure to compare synthetic data from the different approaches, we use a utility metric $\psi$, proposed by \citet{karrFrameworkEvaluatingUtility2006} and extended by \citet{snokeGeneralSpecificUtility2018}. The idea behind this metric is that if a synthetic dataset has a high quality in terms of utility, a classification model cannot distinguish the synthetic samples from real observations well. This means that, ideally $p(x_i \in S_{\mathrm{syn}}) \sim p(x_i \in S_{orig})$, where $S_{\mathrm{syn}}$ is the synthetic dataset and $S_{\mathrm{orig}}$ is the original dataset. Therefore, if we can show the probability of being a member of the synthetic dataset is around \( 0.5 \), we can claim that synthetic and original datasets have similar distributions. Therefore, we combine these two datasets, and add a label variable $y_i$, where $y_i = 1$ if ($x_i \in S_{syn}$) and $y_i = 0$ if ($x_i \in S_{\mathrm{orig}}$). Following this, we apply the Classification and Regression Tree (CART) method to construct a decision tree. We choose the CART model because it excels in scenarios where the original dataset deviates from a normal distribution due to its ability to form decision boundaries in complex, non-linear data spaces. Using this fitted model, we can predict $\widehat{y}_{i}$ for $x_{i=1,\ldots N}$ where $N = n_{\text {syn }} + n_{\text {orig }}$, which is the probability of each observation belonging to synthetic data. The more $\widehat{y}_{i}$ deviates from the ratio of synthetic data size to the merged data size $\left(c=\frac{n_{s y n}}{N}\right)$, the less similar are original and synthetic data. Hence, this utility metric can be measured as shown in Equation ~\ref{eq:utilitymetric}.

\begin{equation} 
    \psi =\frac{1}{N} \sum_{i=1}^{N}\left(\widehat{y}_{i}-c\right)^{2}
    \label{eq:utilitymetric}
\end{equation}

\citealp{snokeGeneralSpecificUtility2018} suggested another metric with the same idea, where the null hypothesis is defined by the CART-based classifier trained on the data with true labels performing as random as for permuted labels, i.e., the original dataset is very similar to the synthetic dataset. Based on several permutations of the labels ($y_i$), the value $\psi_{\mathrm{perm}_j}$ can then be calculated as in Equation ~\ref{eq:utilitymetric}, where $\mathrm{perm}_j$ is the $j$-th permutation. We can then calculate the mean over all iterations $\bar{\psi}$ using Equation ~\ref{eq:utilitymetricmean}. We set the number of permutations to $100$. Then, the final metric is given by Equation ~\ref{eq:utilitymetricratio}.

\begin{equation} 
    \bar{\psi} = \frac{1}{n_{\mathrm{perm}}} \sum_{j=1}^{n_{\mathrm{perm}}} {\psi}_{\mathrm{perm}_{j}}
\label{eq:utilitymetricmean}
\end{equation}

\begin{equation} 
    \psi_{ratio} =\frac{\psi}{\bar{\psi}\vphantom{\frac{\psi}{\psi}}}
    \label{eq:utilitymetricratio}
\end{equation}

In the second approach, we use visual comparisons of marginal distributions. This way, we can check which methods can reconstruct the marginal distributions and which ones and to what extent fail to do so.

 \subsection{Datasets and Results} \label{subsec:datasets}
\subsubsection{Simulation data} \label{subsubsec:sim}
To evaluate our method, we use a published realistic simulation design based on a large breast cancer study \citep{schmoorRandomizedNonrandomizedPatients1996, sauerbreiBuildingMultivariablePrognostic1999}. Specifically, we use the specification published on Zenodo by \cite{zollerModifiedARTStudy2020}. In this simulation study, there is an Exposure variable indicating two different cohorts, i.e., the patients exposed to radiotherapy and non-exposed patients. The outcome of this dataset, denoted as $y$, is defined as having 5-year progression-free survival. The sample size equals 2,500, and there are 21 variables, where 12 are binary, and the rest are continuous variables. Since this simulation data is designed to have real-world distributions, it contains moderate to highly skewed variables. We modify the dataset to additionally include a variable with a bimodal distribution. For this, we generate a bimodal distribution based on the exposure variable by sampling from $N(0,1)$ for $E=0$ and sampling from $N(4,1)$ for $E=1$. The obtained bimodal distribution is also attractive for evaluating our proposed approach because this distribution is not symmetric due to the imbalanced distribution of the exposure variable. On the other hand, we pick the mean of two normal distributions such that the modes are not very far and have overlaps, making it harder for the VAE to imitate the data. As explained above, we optimize the parameters of the pre-transformations and then train the VAE. Table ~\ref{tab:cart-sim} shows the quantitative comparisons explained above. The proposed approach has the best performance based on both of these criteria, followed by the NORTA-J approach. The decision trees are fitted with a minimum leaf size of 20 and a maximum depth of 25.

\begin{figure}[htb]
\begin{center}
\includegraphics[width=0.9\textwidth]{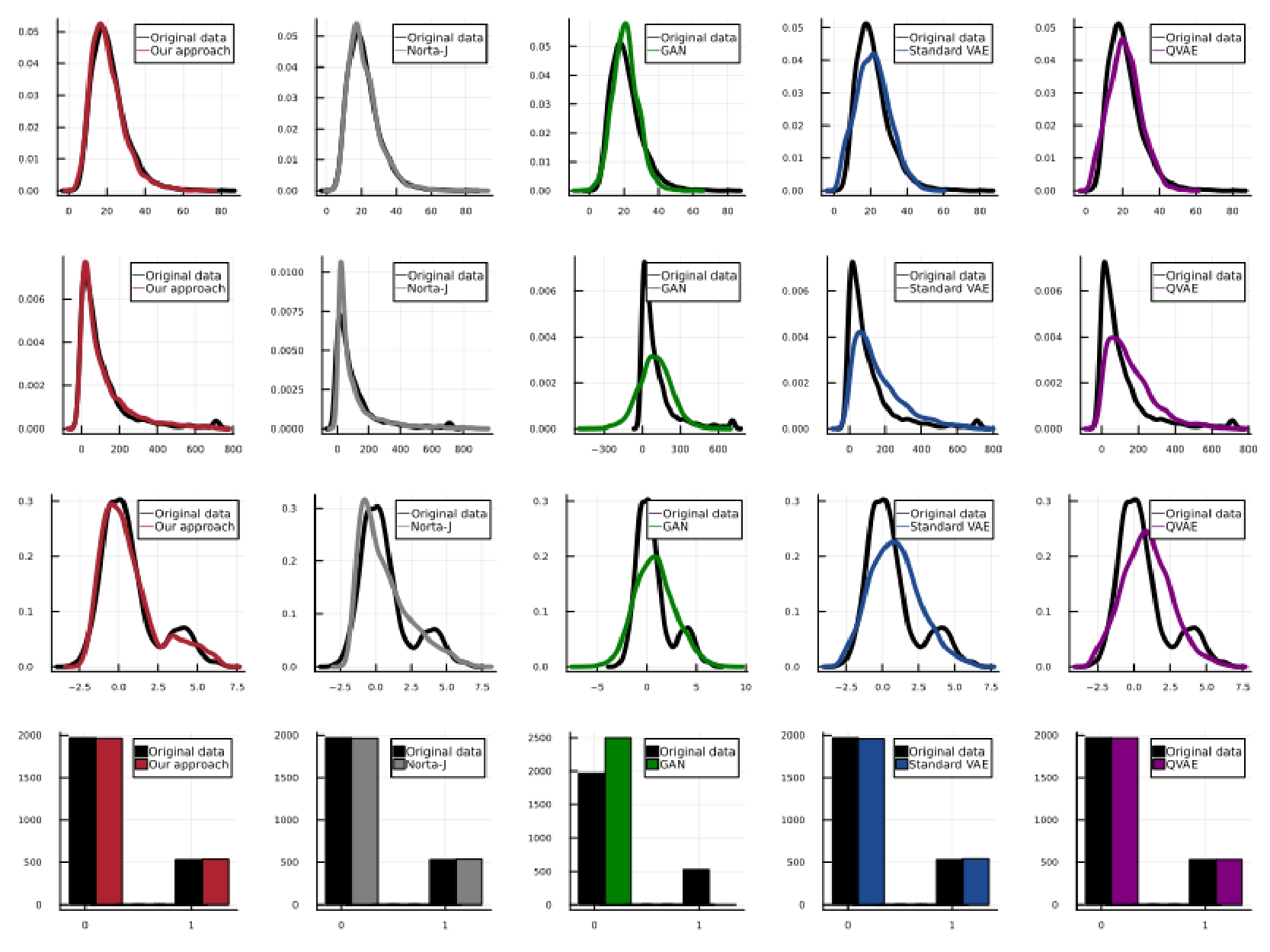}
\caption{Visual comparisons of marginal distributions in synthetic dataset generated by different methods. In this figure, we show four different variables with different types of distributions. The first row shows the slightly skewed variables, the second is the severely skewed variable, and the third row shows the bimodal variable. The fourth row shows a binary variable. In the columns, different methods are illustrated.}
\label{fig:sim_marginal}
\end{center}
\end{figure}

\begin{table}[htb]
\begin{center}
\caption{Comparison of synthetic data generated from simulation data, evaluated by two utility metrics.}
\begin{tabular}{llllll}
\hline \multirow{2}{*}{Metric} & \multicolumn{5}{c}{Methods} \\[3pt]
\cline { 2 - 6 } & Our method & VAE & QVAE & GAN & NORTA-J \\[3pt]
\hline $\psi$ & $\textbf{0.08}$ & $0.12$ & $0.13$ & $0.18$ & $0.10$ \\[3pt]
$\psi_{ratio}$ & $\textbf{1.90}$ & $2.74$ & $2.74$ & $4.18$ & $2.17$ \\[3pt]
\hline
\end{tabular}
\label{tab:cart-sim}
\end{center}
\end{table}

In addition to quantitative comparisons, Figure ~\ref{fig:sim_marginal} shows the visual comparisons of the marginal density diagrams and histograms of selected variables. Here, we show three exemplary continuous variables (slightly skewed, severely skewed, and bimodal) generated by different methods in comparison to the original data and the histograms of a binary variable. As illustrated, our method can generate both slight and severe skewness in the data, i.e., first and second rows. This is when the other variations of the VAE cannot reconstruct the skewness as realistically as the original data, and the trained GAN also fails to do so. Norta-J can perfectly reconstruct the slight skewness, but when it comes to severe skewness, our approach outperforms it with respect to the mode and range of data. In addition, our approach is the only one that can reconstruct the bimodality. For GANs, the problem of mode collapse makes synthetic data generation with bimodality more complicated, in particular when there are different unknown sub-groups. It is worth mentioning that we use different sets of hyperparameters for the deep learning-based approaches, and we pick the most robust results. 

\subsubsection{Real data} \label{subsubsec:real}
For a real data evaluation, we consider the IST dataset, which originates from a large international multi-center clinical trial for stroke patients \citep{sandercockInternationalStrokeTrial2011} and was also used in our other work in \cite{bonofiglioRecoveryOriginalIndividual2020}. Specifically, we use a subset of variables, including randomization variables, i.e., conscious state (RCONSC $=$ drowsy, unconscious or alert), the delay between stroke and randomization (RDELAY in hours), gender (SEX $=$ male/female), AGE, RSLEEP (symptoms noted on waking yes/no), atrial fibrillation (RATRIAL$=$ yes/no), CT before randomization (RCT $=$ yes/no), infarct visible on CT (RVISINF $=$ yes/no), heparin with 24 hours prior to randomization (RHEP24 $=$ yes/no), aspirin with three days prior to randomization (RASP3 $=$ yes/no), systolic blood pressure (RSBP), trial aspirin allocated (RXASP $=$ yes/no, trial heparin allocated (RXHEP $=$ yes/no). We exclude the other randomization variables because of the high proportion of missing values. In addition to this, we used FDEAD, i.e., the outcome defined as dead at six-month follow-up. We also added COUNTRY and derived the REGION (EU-EAST, EU-NORTH, EU-WEST, and EU-SOUTH) from that, to have labels for known sub-groups in the data for using the propensity score-based approach. Excluding the individuals with missing values and those in EU-WEST and EU-SOUTH, we create a rather small dataset with 2,668 records. Among these features, blood pressure, age, and the RDELAY are continuous, the level of consciousness is categorical (with three different values), and the rest are binary. The variable RDELAY has bimodality. We change the level of consciousness to two binary variables (RCONSC1 $=$ drowsy/alert and RCONSC2 $=$ unconscious/ alert) as in \cite{bonofiglioRecoveryOriginalIndividual2020}. We follow the same steps as the steps in the simulation data application, optimizing the parameters of the pre-transformations and then training the VAE. Table ~\ref{tab:cart-ist} shows the quantitative comparisons. For the real data, we see that the group of VAE approaches outperforms the other methods, and the pre-transformation approach has the best results. The decision trees are fitted with a minimum leaf size of 20 and a maximum depth of 25.

\begin{table}[htb]
\begin{center}
\caption{Comparison of synthetic data generated from IST data, evaluated by two utility metrics.}
\begin{tabular}{llllll}
\hline \multirow{2}{*}{Metric} & \multicolumn{5}{c}{Methods} \\ [3pt]
\cline { 2 - 6 } & Our method & VAE & QVAE & GAN & NORTA-J \\ [3pt]
\hline $\psi$ & $\textbf{0.14}$ & $0.15$ & $0.15$ & $0.18$ & $0.25$ \\ [3pt]
$\psi_{ratio}$ & $\textbf{4.40}$ & $4.85$ & $4.97$ & $6.42$ & $5.72$ \\ [3pt]
\hline
\end{tabular}
\label{tab:cart-ist}
\end{center}
\end{table}

\begin{figure}[htb]
\begin{center}
\includegraphics[width=0.9\textwidth]{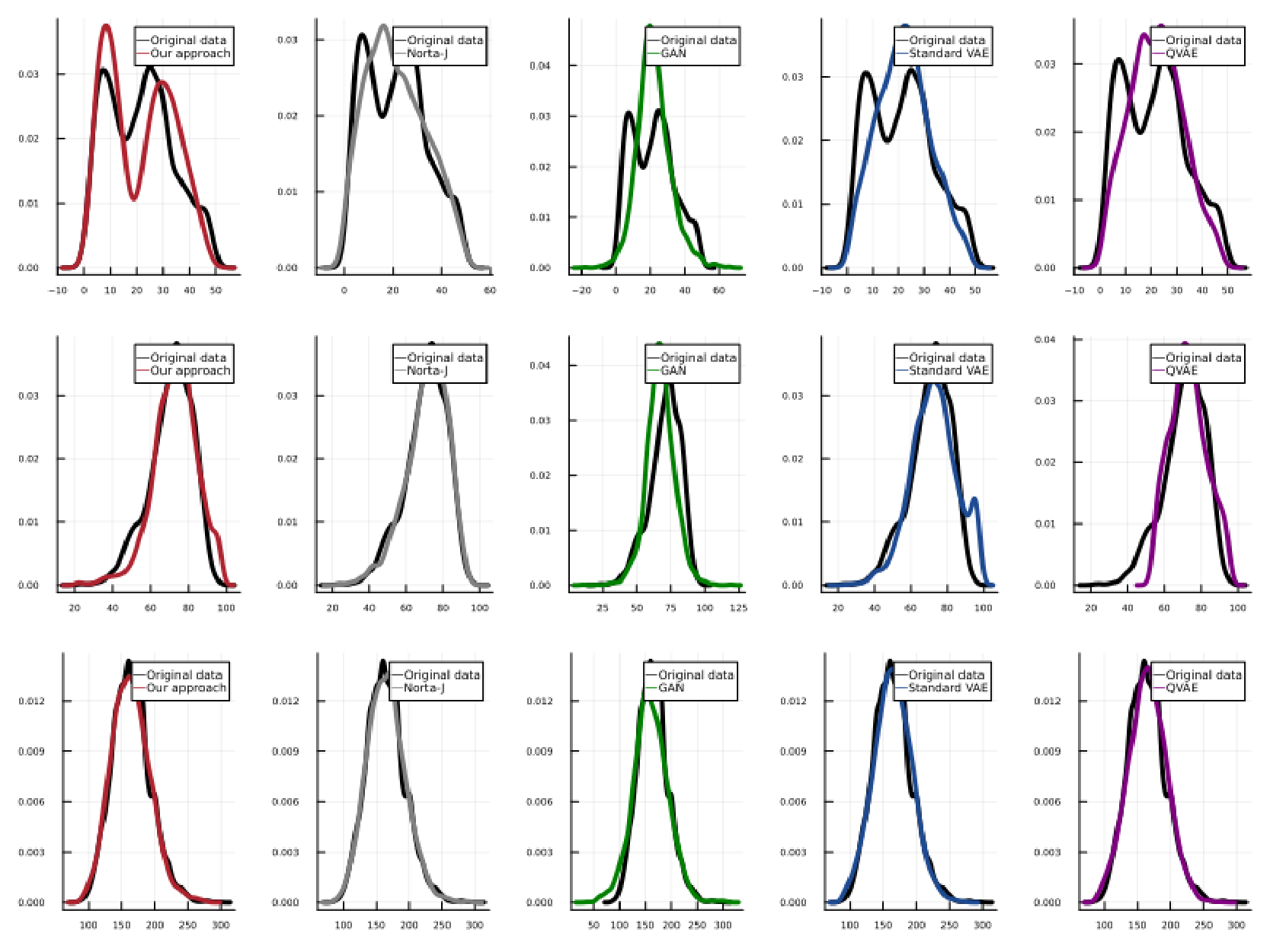}
\caption{Visual comparisons of marginal distributions in synthetic dataset generated by different methods. This figure shows three continuous variables, including a bimodal distribution, i.e., shown in the first row. In the columns, different methods are illustrated.}
\label{fig:propensity_ist_syn}
\end{center}
\end{figure}

In addition to quantitative comparisons, Figure ~\ref{fig:propensity_ist_syn} shows the visual comparisons of the marginal density diagrams. Here, we show three continuous variables, which exist in the dataset by baselines in comparison to the original data. As illustrated in the first row, our method can generate the bimodality of the RDELAY variable in the data in contrast to the other variations of the VAE, which generate an unimodal distribution. We use different sets of hyperparameters for the deep learning-based approaches and pick the most robust results.

\section{{Evaluation of the method for known sub-group structures}} \label{sec:propensity-score-reults}
\subsection{Simulation data}
We start with the simulation design to explore the possibility of integrating propensity scores with the latent representation of VAE. For this, we have a validity check based on our previous study \citep{zollerAutomaticVariableSelection2020}. Since variable selection is one of the challenging steps of propensity score model building, in the previous study, we investigated the simulation design to see whether the variables should be selected in relation to exposure alone or both exposure and outcome. In this use case, we concluded that selecting variables based on exposure for this breast cancer-based simulation study gives more reliable results for estimating the propensity score. To inspect whether the latent representation of the VAE also captures these patterns, we overlay a heat map based on the propensity score grid with a scatterplot of the latent representation color-coded by two cohorts (exposed and non-exposed). If the color patterns from the propensity scores, calculated with variables related to exposure, align better with the color of data points in the latent space, in comparison to the variable selection considering both exposure and outcome, it would confirm our previous conclusion. This would suggest that our methodology is indeed promising for guided prior sampling. For training the VAE, we excluded the exposure and outcome variables. Moreover, we excluded the variable $x_6$, corresponding to the progesterone receptor status, because the data is simulated such that $x_6$ is related to both exposure and outcome and can be approximated by other variables. So, this way, we can have a scenario that has an unmeasured confounder. Then, we can investigate whether the propensity score-based values match the latent structure and check if the latent structure corresponds to the mentioned results in \cite{zollerAutomaticVariableSelection2020}.

\begin{figure}[htb]
\begin{center}
\includegraphics[width=1.0\textwidth]{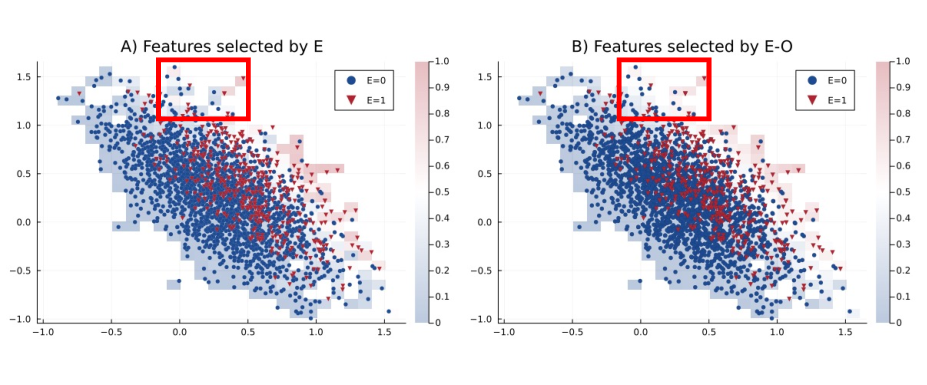}
\caption{The latent representation of simulation design removing the confounding variable, $x_6$, extracted by VAE. In this plot, blue dots represent the non-exposed cohort, and the red triangles represent the exposed cohort. A) shows the heat map color-coded based on the propensity score, which is calculated by a selection of variables related to exposure, and B) the heat map color-coded based on the propensity score, which is calculated by a selection of variables related to both exposure and outcome. The area outlined by the red square shows the differences between the two variable selection methods. }
\label{fig:propensity_sim}
\end{center}
\end{figure}

Then, we use logistic regression on the original variables to predict the exposure and selected variables related to exposure if their $p$-value was smaller than $0.05$. Then, we selected variables related to the outcome by applying the logistic regression on the original data for predicting the outcome and including variables with a $p$-value smaller than $0.05$. Then, we fitted two models. The first model, called the $E$ model, is fitted using the first variable selection. The second model is the $E-O$ model, which uses only the variables that are selected for both exposure and outcome. In Figure ~\ref{fig:propensity_sim}, we see that regardless of the variable selection method, the latent structure matches the propensity score-based values reflected in the colored grid behind the latent representation. Moreover, the area outlined by the red square shows that as this area has more blue data points, i.e., representing the non-exposed individuals, the propensity score model, which generates more blue grid cells would be the better approach. Therefore, having the model with variables related to exposure is a better match with the latent representation generated by VAE. Consequently, since the results align with the previous study, we can conclude that combining propensity score regression with VAEs can be a promising sampling guide for VAEs.

\subsection{Real data}

\begin{figure}[t]

\begin{center}
\includegraphics[width=1.0\textwidth]{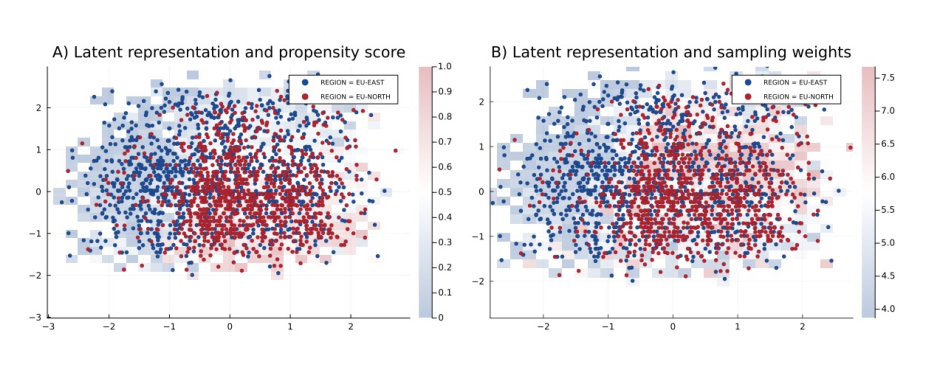}
\caption{The latent representation of IST data, extracted by VAE. In both A and B, the blue dots represent the observations that belong to the region EU-EAST, and the red dots represent the observations that belong to EU-NORTH. In A, the heat map is color-coded based on the averaged propensity score, which is calculated by variable selection related to the region. In B, the heat map is color-coded based on the calculated weights for prior sampling when the target scenario is to remove the systematic differences.}
\label{fig:propensity_ist}
\end{center}
\end{figure}

In the real data example, the sub-groups are related to sub-groups in the dataset are related to the moderating variable of region membership, since it effects in different ways, e.g., variables related to the healthcare system or population-specific characteristics. Therefore, we use the REGION variable for the propensity score, fitting the logistic regression on original values predicting the REGION (EU-NORTH $= 1$ and EU-EAST $= 0$), and we select variables according to $p$-value with the cutoff $\alpha$ set to $0.05$. Then, using the weighting approach for generating individuals common for both sub-groups from Equation ~\ref{eq:commonclassindweighting}, we calculate the weights for the weighted sampling from the prior explained in Section ~\ref{subsec:psweightedsampling}. Getting the latent representation from the trained VAE, explained in ~\ref{sec:marginal-results}, and overlaying the propensity score heat map and weight heat map, we obtain Figure ~\ref{fig:propensity_ist}. The left plot in the figure confirms the feasibility of combining propensity score regression with the latent representation of VAE, as the areas with a majority of red dots correspond to the red grid cells. In the right plot, the red grid cells correspond to the areas with larger weights, i.e., with less systematic differences between the two sub-groups, and the blue grid cells correspond to the areas with sub-group-specific characteristics.

\begin{figure}[t]
\begin{center}
\includegraphics[width=0.9\textwidth]{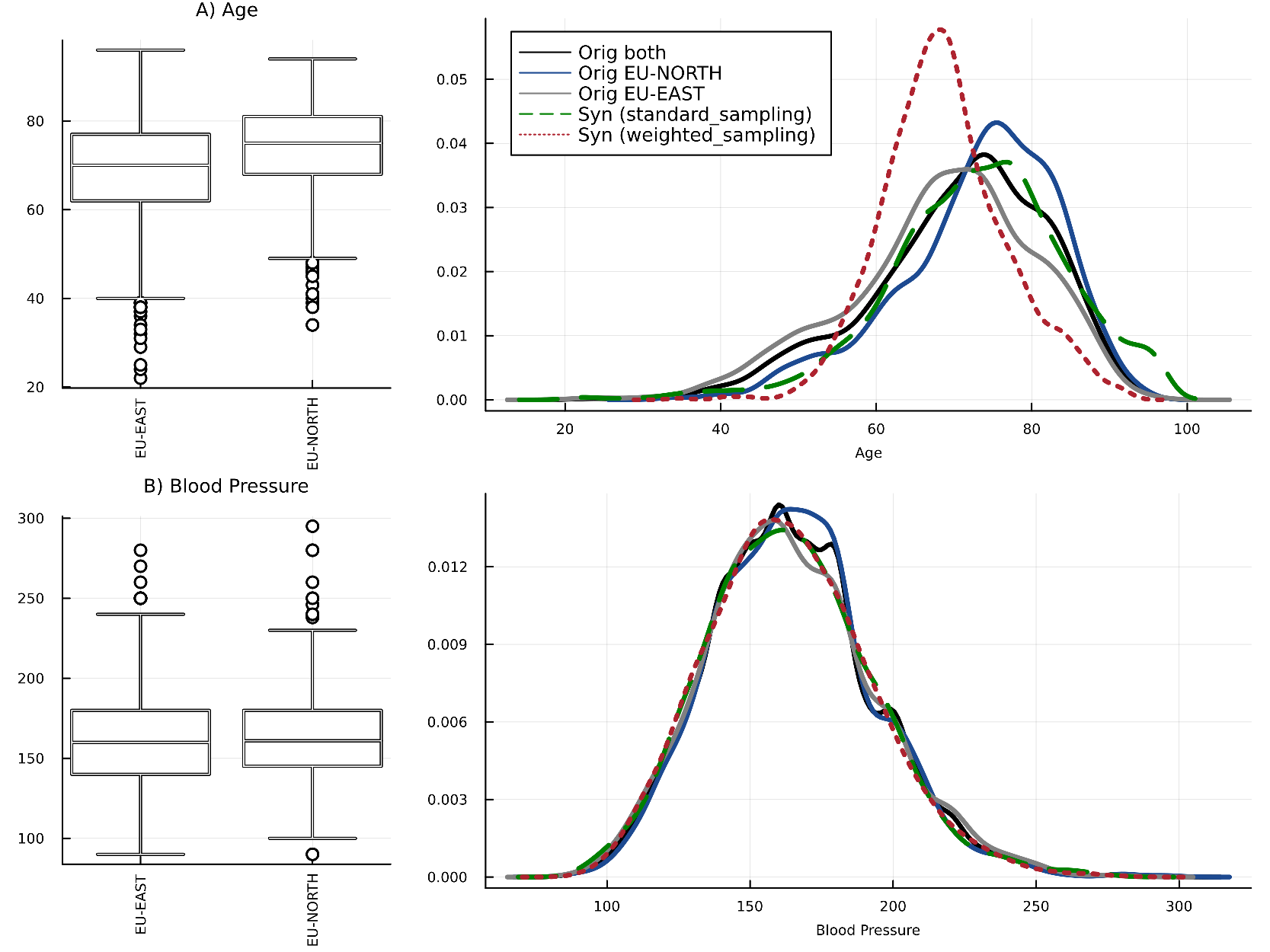}
\caption{Visual comparisons of marginal distributions in synthetic dataset generated by different methods. This figure shows three continuous variables, including a bimodal distribution, i.e., shown in the first row. In the columns, different methods are illustrated.}
\label{fig:propensity_ist_marginal}
\end{center}
\end{figure}

To investigate the impact of our approach, we compare marginal distributions from the two populations and generate data using standard and weighted sampling approaches. For this, we choose blood pressure, which has a similar distribution across the regions, and age, which is differently distributed, e.g., with a  higher age of stroke in the EU-NORTH population. So, in this specific scenario, removing systematic differences means that in the synthetic data, we should not have a very high frequency of older individuals. The red dashed line in Figure ~\ref{fig:propensity_ist_marginal}.B for the blood pressure variable shows that our approach recognizes no systematic differences for this variable. Therefore, the generated data has the same marginal distributions in both sub-groups. Still, when it comes to age, the marginal distribution is completely different (red dashed line in Figure~\ref{fig:propensity_ist_marginal}.A), having a higher peak but almost similar mode to EU-EAST. The explanation for this is that because of differences in the population or in the healthcare system, EU-NORTH has a different underlying distribution. Getting back to the latent structure in Figure ~\ref{fig:propensity_ist_syn}.B, the areas with red grid cells, i.e., with smaller weights, have a higher concentration of EU-NORTH members. Therefore, we have fewer samples from those areas, and can ensure that we do not have, e.g., many individuals with stroke age of \( 80 \) and generate a population that is on average younger than EU-NORTH.

For this result, we set the threshold for a zero weight $\delta$ from Equation~\eqref{eq:commonclassindweighting} to $0.1$. Lower values, $\delta \approx 0$ are suitable for the scenarios where we are interested in preferentially sampling from the areas that have a rather similar group membership probability, while for higher values of $\delta$, we include samples which may be more common to one sub-group but still can be found in other sub-group as well. 

Overall, the results show that the weighted sampling approach is helpful when dealing with known sub-groups.

\section{Discussion} \label{discussion}
Variational autoencoders (VAEs) have shown promising results for generating image data, which is often evaluated based on the overall visual impression without analyzing individual pixel distributions. In contrast, synthetic clinical cohort data has different requirements,  as heterogeneity is often a critical characteristic. Heterogeneity may be due to known sub-groups, e.g., reflecting different study sites, or may be unknown and just be reflected in marginal distributions. We investigated whether combining deep learning and classical statistical approaches --- specifically pre-transformations for addressing heterogeneity reflected in bimodal or skewed distributions and propensity score regression for addressing known sub-groups --- might be useful for synthetic data generation.

We used a realistic simulation based on a breast cancer study and showed that the proposed approach can reconstruct the complex marginal distributions, thus preserving the unknown sub-group structure. To see if propensity score estimation on the original data space can complement the VAE approach, we considered visualization in the latent VAE representation and found that propesńsity scores add complementary information. We illustrated the approach with a real dataset from an international stroke trial. The results show that our approach can reconstruct the more complicated marginal distributions, such as bimodal ones, even in the presence of different categorical/binary variables. We could obtain a latent representation that was useful for subsequent propensity score-guided sampling. Thus, extremes of sub-groups could be avoided in synthetic data. 

Certainly, the proposed approach cannot address all potential types of heterogeneity, as we focused on bimodal and skewed marginal distributions, i.e., there might be other complex distributions that our approach cannot recover completely. Yet, these two are the most common marginal distributions in biomedical settings. Regarding the known sub-groups, we so far have not optimized the propensity score model, despite known challenges in model building \cite{austinPerformanceDifferentPropensity2013}. Consequently, the proposed approach could probably be improved, e.g., by more closely investigating variable selection approaches for constructing the propensity score. 

To summarize, the proposed approach illustrates that it can be useful to complement VAEs with more classical statistical modeling approaches for addressing heterogeneity when generating synthetic data. This can more generally pave the way for high-quality synthetic clinical cohort data in presence of sub-groups.

\section*{Acknowledgments}
The work of MH, DZ, and HB was funded by the Deutsche Forschungsgemeinschaft (DFG,  German Research Foundation) -- Project-ID 499552394 -- SFB 1597.

\section*{Conflict of Interest}

\noindent {\it{The authors have declared no conflict of interest.}}


\bibliographystyle{abbrvnat}  
\bibliography{PTVAE}

\end{document}